\documentclass[11pt,a4paper]{article}
\usepackage{acl2018}
\usepackage{times}
\usepackage{latexsym}
\usepackage{graphicx}
\usepackage{algorithm,algorithmic}
\usepackage{amsmath}
\usepackage{amssymb}
\usepackage{framed}

\usepackage{scrextend}
\usepackage{textcomp}

\usepackage{url}

\aclfinalcopy 

\title{Embedding Grammars}

\author{David Wingate \\
  {\tt wingated@cs.byu.edu} \\\And
  William Myers \\
  {\tt mwilliammyers@byu.edu} \\
  \AND
  Nancy Fulda \\
  {\tt nfulda@byu.edu} \\\And
  Tyler Etchart \\
  {\tt tyler.etchart@byu.edu} \\}

\date{}

\def\dstt#1{\begin{small}\texttt{#1}\end{small}}

\begin{document}

\maketitle

%
%

\begin{abstract}
Classic grammars and regular expressions can be used for a variety of purposes, including parsing, intent detection, and matching.  However, the comparisons are performed at a \emph{structural} level, with constituent elements (words or characters) matched exactly.  Recent advances in word embeddings show that semantically related words share common features in a vector-space representation, suggesting the possibility of a hybrid grammar and word embedding. In this paper, we blend the structure of standard context-free grammars with the semantic generalization capabilities of word embeddings to create hybrid semantic grammars.  These semantic grammars generalize the specific terminals used by the programmer to other words and phrases with related meanings, allowing the construction of compact grammars that match an entire region of the vector space rather than matching specific elements.
\end{abstract}




%
%

\section{Introduction}

Consider the problem of intent detection for natural language processing: to estimate intent, a natural language sentence must be analyzed to determine the underlying goal of the speaker.  Ideally, such analysis would be insensitive to the specific words and phrases used, implying the need for a general semantic metric for equivalence.


One way to approach this problem would be to craft a simple rule---say, a regular expression---that describes a set of semantic equivalences.  A regex for detecting an intent to deliver a compliment might resemble the following:


\noindent
\begin{framed}
\begin{center}
  \dstt{(i think)? (you're|you are)\\ (beautiful|gorgeous|cute)}
\end{center}
\end{framed}

\noindent This would match sentences such as \dstt{I think you are beautiful}, \dstt{you are cute}, and \dstt{you're gorgeous}.
However, the ability of this regex to match input sentences is limited by the mental lexicon of the regex designer. What the designer \emph{intended} was to match any phrase that could be construed as a compliment regarding the auditor's general physical appearance, but the produced regex provides only a compact set of examples.  The regex does not include all possible synonyms for \dstt{beautiful}, and so it would not, for example, match the sentence \dstt{I think you are lovely}.  The power of grammars and regular expressions lie in the compact and efficient representation of combinatorially large sets of sentences, but it is always assumed that terminal matches must be exact.  Here, we relax that assumption.

This paper sketches out how to combine the strengths of context-free grammars with the generalization capability of word embeddings to create semantic regular expressions. The grammar provides the coarse overall structure, but individual tokens serve primarily as exemplars that define a region in a word embedding feature space. Any token whose feature representation lies within this region is considered a match. The result is semantic, rather than structural expression evaluation.



%
%

\section{Background}
\label{sec:background}

Word embeddings \cite{Bengio2003, Pennington2014Glove, Mikolov2013EfficientEstimation, bojanowski2016fasttext}
represent the meaning of a word as a vector of learned features with interesting linear properties. They have not only been used to evaluate analogies
\cite{Mikolov2013LinguisticRegularities}\cite{nematzadeh2017},
but also for document retrieval\cite{Georgios2016}, affordance detection \cite{Fulda2017Rock}, and the generation of sentence-level embeddings \cite{Kiros2015skipthought}. Multimodal embedding algorithms inspired by language models have also
been used for text and image alignment \cite{Frome2013, Weston2011, GarciaGasulla2015}, 

Algorithms for learning word embeddings typically assume that words in similar contexts should have feature representations that are close to one another, implying that semantically related words will be proximately located. Our work utilizes 300-dimensional fastText embeddings trained on a Wikipedia corpus \cite{bojanowski2016fasttext} which, like word2vec \cite{Mikolov2013EfficientEstimation}, GLoVE \cite{Pennington2014Glove}, and related vector sets, were trained  using a neural network tasked with predicting local context. Unlike previous methods, fastText utilizes subword information to speed the training process and generate compositional embeddings for previously unseen words.


\begin{figure*}
\centering
\includegraphics[width=\textwidth]{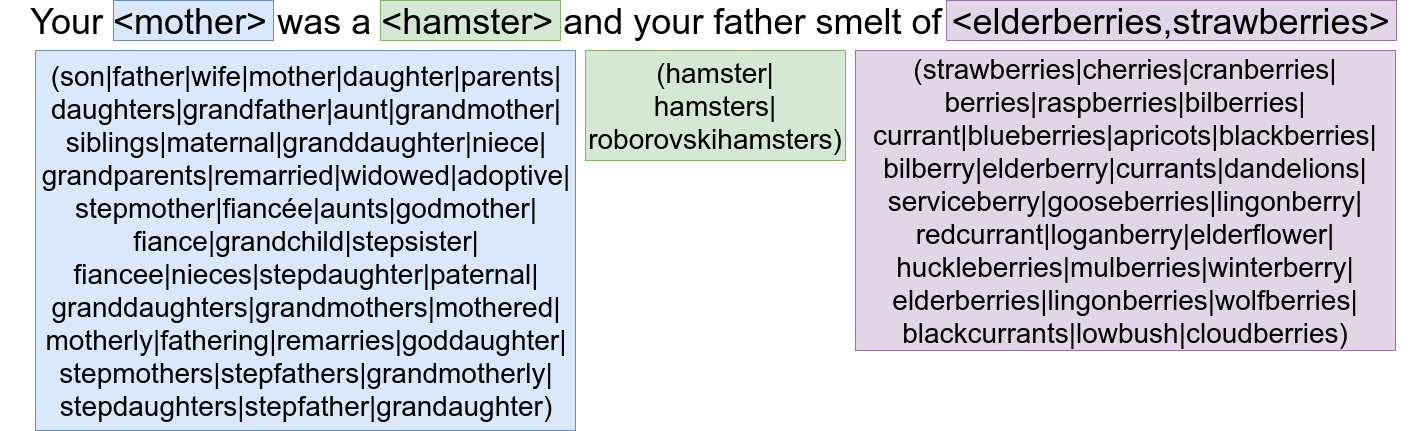}
\vskip -0.15in 
\caption{Sample regexv expansions obtained using the 500,000 most common tokens in the English-language fastText corpus. Matches were selected using the neighbor expansion method with $\epsilon$=0.35.
}
\label{fig:elderberries}
\end{figure*}

Known weaknesses of word embeddings include susceptibility to triangle inequalities and breakdowns in symmetry \cite{nematzadeh2017} as well as an inability to represent different semantic meanings of the same word. Researchers are currently exploring disambiguation methods to distinguish between multiple senses of a given word \cite{reisinger2010, Huang2012} or to combine the strengths of several models \cite{ghannay2016}.


%
%

\section{Embedding Grammars}

A \emph{context-free grammar} (CFG) is defined as a tuple $(N,W,R,S)$ where $N$ is a finite set of nonterminal symbols, $W$ is a finite set of terminal symbols, $R$ is a finite set of productions of the form $A\rightarrow B$ with $A\in N$, $B\in \left( W \cup N \right)*$ (where $*$ is the Kleene star), and where $S\in N$ is a distinguished start symbol.
The language generated by the grammar is the yield of the terminal strings generated with all possible (recursive) substitutions of right-hand side nonterminals with left-hand side nonterminals.

An embedding grammar generalizes classic CFGs: instead of considering terminals $W$ to be ``grounded'' in discrete tokens, we instead consider terminals to be grounded in a well-defined region of a feature space.  We therefore replace the set of symbols $W$ with a set of subregions $W_e$, where each $r\in W_e$ is defined as some $r\subset\mathbb{R}^d$, where $d$ is the dimension of the feature space.

To connect tokens and the grammar, we require a \emph{word embedding} using one of the methods described in Sec.~\ref{sec:background} that defines a function $f$ that maps tokens $w\in W$ to points in $\mathbb{R}^d$.  We then say that a production $A\rightarrow B$ is valid if $B\in W_e, w\in W$, and $f(w) \in B$.
 
This naturally raises the question: how should the subregions be specified?  We adopt an exemplar based approach: the grammar designer specifies the subregions indirectly by giving several exemplars, which are then used in conjunction with the word embedding database to automatically generate the subregion.

Figure \ref{fig:elderberries} illustrates the idea for a simple BNF grammar given by: 
\begin{small}
\begin{verbatim}
  S ::= Your REL was a PET and your
        father smelt of BERRIES
  REL ::= mother
  PET ::= hamster
  BERRIES ::= elderberries
\end{verbatim}
\end{small}
where nonterminals are given in all capital letters.  Naively, this simple grammar would yield only the single sentence ``Your mother was a hamster and your father smelt of elderberries'' \cite{gilliam75}.  However, using embedding grammars, we can interpret the production \dstt{REL ::= elderberries} as an exemplar, and define the subregion $r$ as (for example) an epsilon ball centered at $f(\mathrm{elderberries})$.  This naturally generalizes the yield of the grammar to a wide variety of familial relationships, pets, and berries.

%
%

\subsection{Subspace region definition}
Subspace definition is independent of the overall grammar structure, and different methods can be used as needed. Here, we outline four possible approaches to subspace definition. (In practical application, we found the neighbor expansion method to be most effective, and use it in our examples throughout the paper.) 

\textbf{Bounding Box:} A naive subspace can be defined by taking a bounding box of the exemplars. As one might expect, this method does not work particularly well, and it also cannot be applied to productions consisting of only a single exemplar. \textbf{Epsilon ball:} In this approach, the centroid of the exemplar set is taken, and all vectors whose cosine distance is less than a predefined value $\epsilon$ from the centroid are considered to fulfill the production rule.  \textbf{Covariance subspace:} We define a Mahalanobis distance metric based on the empirical (possibly regularized) covariance of the exemplars. We expected this method to outperform the epsilon ball, since it takes into account the idea that certain components of the vector space have a greater influence on word semantics. In reality, the opposite was true. \textbf{Neighbor expansion:} We found this method to be the most successful. In this case, each exemplar in a given production defines a subspace consisting of all vectors whose cosine distance to said exemplar is less than $\epsilon$. Thus, a vector which is within the epsilon ball of $any$ exemplar is considered to match the production rule. This many-bubbles approach results in close matches for the entire set of exemplars.


%
%

\section{A Python Implementation}


To test our ideas, we experimented with a simple integration of embeddings and Python's regular expression package (\dstt{re}).  The Python regex package \cite{regexgithub2017} provides a concise notation for describing sets of character strings, allowing complex string matching based on search patterns entered by a human designer. In Python, regex uses a backtracking algorithm rather than translating the regex into an equivalent deterministic finite automaton (DFA) or non-deterministic finite automaton (NFA).
We took advantage of \dstt{re}'s dynamic nature by defining and calling a wrapper function which applies regexv before calling each of \dstt{re}'s public API methods.





\section{Experiments}
Our experiments used 300-dimensional FastText vectors trained on Wikipedia. A coarse parameter search suggests that $\epsilon$ $<$ 0.5 represents a strong semantic similarity. However, because polar opposites (e.g. beautiful/ugly) appear in highly similar contexts and tend to fall within this range of each other, we adopted a stricter match threshold of $\epsilon$=0.35. The neighbor expansion method discussed in Section 3.1 was used for all experiments.


\subsection{Word extensions}
Sample word expansions are shown in Figure 1. The common and socially-integral term `mother' (777th most common token) produces a wide range of matches, whereas the less common word `hamster' (33651st most common token) defines a relatively sparse subspace.

Note that grammatically incorrect matches such as `widowed' or `motherly' are unlikely to cause poor system performance, as human-generated text will almost never contain syntactically implausible phrases like ``Your widowed was a hamster''. The same holds true for unusual and entertaining words such as `roborovskihamsters'. The most problematic error modes involve words that are polar opposites: hot/cold, happy/sad, high/low, because they tend to appear in similar contexts. 

Multi-word phrase matching was accomplished by treating each input phrase as the vector average of its individual words, an approach that to our surprise functioned quite well. For example, the vectors for `clever' and `man', when averaged together, produce a vector that is highly similar to the vector for `genius'. Further examples are shown in Table 1.

\begin{table}
\begin{center}
 \begin{tabular}{|| c | c | c ||} 
 \hline
 \textbf{match phrase} & \textbf{exemplar} & \textbf{cos distance} \\
 \hline
 \hline
 clever man & genius & .3934 \\
 clever car & genius & .5776 \\
 smart boy & genius & .4822 \\
 smart car & genius & .6859 \\
 fast car & genius & .7700 \\
 smart car & tesla & .5758 \\
 smart car & porsche & .4609 \\
 \hline
 a guy who & & \\ 
 knows everything & genius & .5060 \\
 \hline
 a guy who & & \\
 knows everything & porsche & .7767 \\
\hline
\end{tabular}
\end{center}
\caption{Cosine distance between single-word vectors and compositional vectors created by averaging the elements of a multi-word phrase. The resulting distances correlate with many human estimations of semantic similarity.}
\end{table}

\subsection{Intent matching}

\vskip 0.1in
\begin{figure*}
\fbox{
  \parbox{\linewidth}{
      \textbf{Regex String: ``I would like to call a \dstt{[taxi]}''}
      \\
      \textbf{regex}: ``I would like to call a (taxi\textbar{}car)''
      \begin{addmargin}{1em}
        \textit{matches}:  ``I  would  like to call a taxi'', ``I would like to call a car''
      \end{addmargin}
      \textbf{regexv}: ``I would like to call a $\langle{}$taxi,car$\rangle{}$''\\
      \textbf{post expansion}: ``I would like to call a (taxi\textbar{}taxis\textbar{}minibus\textbar{}taxicab\textbar{}taxicabs\textbar{}car\textbar{}cars\textbar{}vehicle\textbar{}driver\textbar{}\\driving\textbar{}truck\textbar{}automobile\textbar{}suv\textbar{}roadster\textbar{}motorbike\textbar{}racecar\textbar{}minivan\textbar{}motorcar\textbar{}racecars)''
      \begin{addmargin}{1em}
        \textit{matches}:  ``I would like to call a minibus'', ``I would like to call a taxicab'', etc.
      \end{addmargin}
  }
}
\caption{Comparison between regex and regexv. Here, \dstt{[taxi]} is shorthand for the intent that would include all words that are synonymous with taxi.}
\label{fig:taxi}
\end{figure*}


Figure \ref{fig:taxi} illustrates another example of intent matching.
If the string one expects is something like ``I would like to call a taxi'', then people could also say ``I would like to call a car'' and intend the same directive. Typical grammars might define a pattern, ``I would like to call a (taxi\textbar{}car)'', in order to match these semantically equivalent sentences. The problem is that the developer must enumerate all the possible words that could mean the same thing as ``taxi'' in this pattern. Regexv solves this issue by allowing a user to define intent words, then automatically expands these to a large list of words that have the same semantic meaning.
Figure \ref{fig:taxi} shows how regexv to expands two terms to an entire list, including taxi, minibus, taxicab, car, vehicle, driver, racecar, etc.



\section{Learning and Extensions to PCFGs}

The fact that terminals in embedding grammars are defined as regions in a continuous feature space suggest a natural extension to probabilistic context-free grammars (PCFGs).  A PCFG is a tuple $(N,W,R,S,\theta)$, where $(N,W,R,S)$ are defined as for a normal CFG, but where each production $A\rightarrow B$ is annotated with a probability $\theta_{A \rightarrow B}$, st $\sum_{A\rightarrow B\in R_A} \theta_{A\rightarrow B}=1$, where $R_A$ is the subset of productions in $R$ with left-hand side $A$.  Informally, $\theta_{A\rightarrow B}$ is the probability of expanding the nonterminal $A$ using the rule $A\rightarrow B$.  This naturally defines a probabilistic generative model, and can be used to (for example) rank parses of ambiguous sentences according to some prior.

Embedding grammars offer an additional generalization: instead of defining matches as hard regions in $R^d$, soft matches could be defined using (eg) Gaussians; this suggests fully probabilistic learning algorithms that could take advantage of soft matches and the continuous nature of word embeddings.  When combined with, for example, recent probabilistic grammars such as adaptor grammars \cite{johnson2007adaptor} and fragment grammars \cite{o2009fragment} that combine Bayesian nonparametrics and PCFGs, this could potentially result in efficient grammar learning algorithms: learnt grammars could be more compact  because of the natural generalization inherent in the word embeddings. 

\section{Conclusion}

Word embeddings can be combined with classic grammars to enable semantic matching of regular expressions. 
This is advantageous because it reduces human cognitive effort, allowing grammar designers to say what they want more compactly. 

Importantly, the matching power of the expression is not limited by the designer's lexicon, but can include words that the designer is not familiar with. Recent advances in word embeddings, including fastText, provide for generalization to previously unseen words based on subword structure.


We have only scratched the surface of the possibilities of embedding grammars.  Extensions to PCFGs and the advantages of embeddings vis-a-vis grammar learning are exciting directions for future work.





\bibliography{acl2018}



\end{document}